\documentclass[11pt,a4paper]{article}
\usepackage[hyperref]{emnlp2018}
\usepackage{times}
\usepackage{latexsym}
\usepackage{graphicx}
\graphicspath{ {./images/} }
\usepackage{multirow, makecell}
\usepackage{tabularx}
\newcolumntype{Y}{>{\centering\arraybackslash}X}
\newcolumntype{P}[1]{>{\centering\arraybackslash}p{#1}}

\usepackage{url}

\aclfinalcopy % Uncomment this line for the final submission

\title{A Byte-sized Approach to Named Entity Recognition}

\author{Emily Sheng \\
  USC Information Sciences Institute \\
  Marina del Rey, CA \\
  {\tt ewsheng@isi.edu} \\\And
  Prem Natarajan \\
  USC Information Sciences Institute \\
  Marina del Rey, CA \\
  {\tt pnataraj@isi.edu} \\}

\date{}

\begin{document}
\maketitle
\begin{abstract}
  In biomedical literature, it is common for entity boundaries to not align with word boundaries. Therefore, effective identification of entity spans requires approaches capable of considering tokens that are smaller than words. We introduce a novel, subword approach for named entity recognition (NER) that uses byte-pair encodings (BPE) in combination with convolutional and recurrent neural networks to produce byte-level tags of entities.  We present experimental results on several standard biomedical datasets, namely the BioCreative VI Bio-ID, JNLPBA, and GENETAG datasets. We demonstrate competitive performance while bypassing the specialized domain expertise needed to create biomedical text tokenization rules.\footnote{https://github.com/ewsheng/byteNER}

%eschewing manually crafted pre- and post-processing rules to improve performance. \footnote{Code will be made available}
\end{abstract}

\section{Introduction}
While NER tasks across domains share similar problems of ambiguous abbreviations, homonyms, and other entity variations, the domain of biomedical text poses some unique challenges. While, in principle, there is a known set of biomedical entities (e.g., all known proteins), there is a surprising amount of variation for any given entity. For example, \textit{PPCA}, \textit{C4 PEPC}, \textit{C4 PEPCase}, and \textit{Photosynthetic PEPCase} all refer to the same entity. Additionally, certain entities such as proteins and genes can naturally span less than a ``word" (e.g., \textit{HA} and \textit{APG12} are separate proteins in \textit{pHA-APG12}). Most state-of-the-art NER methods tag entities at the ``word" level, and rely on pre- or post-processing rules to extract subword entities. Our goal is to develop a subword approach that does not rely on ad hoc processing steps.

To that end, we introduce a novel subword approach to identifying named entities. Our decision to work with input features and output tags at the byte level instead of the character level is because biomedical datasets typically provide byte offset annotations; however, our methods may also be applied to character-level models. In this paper, we refer to ``subword models'' as models that take as input a sequence of subwords (e.g., bytes) and output a corresponding sequence of subword tags (e.g., one tag per byte). Our focus is the effects of different subword features on identifying named entities in various biomedical NER datasets, which is especially useful for entities that are arguably more naturally annotated at the subword level.

\section{Related Work}

% Traditional approaches to NER include hand-engineered features for Maximum Entropy models \cite{curran2003language}, Conditional Random Fields (CRFs) \cite{mccallum2003early}, and Hidden Markov Models (HMMs) \cite{klein2003named}. 
State-of-the-art neural NER techniques developed in recent years use a combination of neural networks (NN) and Conditional Random Fields (CRFs) to achieve high precision and recall of named entities. These techniques pass word and character embeddings to a bi-directional long short term memory (BLSTM) layer, which may be followed by a CRF layer \cite{ma2016end,lample2016neural,chiu2015named}. 
% Generally, the CRF component and pre-trained word embeddings greatly improve the performance of neural NER models. 
These state-of-the-art techniques have also been successfully applied to biomedical datasets \cite{lyu2017long,gridach2017character}. Although these techniques use ``subword'' features such as character embeddings, these models take as input a sequence of words and output a sequence of word tags, and are thus different from what we refer to as subword models in this paper. We build upon state-of-the-art neural techniques to evaluate models that take subword input features and produce corresponding subword output tags.

Subword models have mostly been developed in the context of multilingual datasets \cite{gillick2015multilingual}, machine translation \cite{sutskever2014sequence}, and processing for character-based languages \cite{misawa2017character}. 
% The way we combine byte features is similar to that of \citet{misawa2017character}. 
\citet{kuru2016charner} develop a model that tags sequences of bytes, though they ultimately relies on word boundaries to determine appropriate tags. \citet{abujabal2018neural} use characters, phonemes, and bytes as subword features, and similarly tag entities at word-level boundaries.

Byte-pair encoding iteratively combines frequent characters to build a ``codebook" of character merge operations \cite{sennrich2015neural,gage1994anew}. 
% Words are segmented into subword tokens in order to better deal with rare and unseen words. 
\citet{verga2018simultaneously} also use BPE as features in their biological relation extraction and NER multi-task model, though they primarily focus on the former task. 
% extraction of relations between genes, chemicals, and diseases. 
% We use BPE-segmented tokens as one source of information for our subword NER models.

\section{Datasets}
The first dataset, BioCreative VI Bio-ID, was introduced for Task 1 of BioCreative VI \cite{arighi2017bioid}, and consists of figure captions with annotations for six entity types.
% from 766 biomedical research papers. This biomedical dataset has 
% with byte offset annotations for \textit{cell types or lines}, \textit{cellular components}, \textit{organisms or species}, \textit{proteins or genes}, \textit{small molecules}, and \textit{tissues or organs}. 
The Bio-ID dataset is the only dataset we experiment on that is annotated with byte offsets and contains raw text that has not been tokenized or converted into ASCII format. The second dataset, JNLPBA, is an annotated set of 2,404 biomedical abstracts \cite{kim2004introduction} with annotations for five entity types.
% Annotations are available for \textit{proteins}, \textit{DNA}, \textit{RNA}, \textit{cell types}, and \textit{cell lines}. 
The third dataset, GENETAG \cite{tanabe2005genetag}, is a collection of 20K sentences from MEDLINE that are annotated with \textit{proteins/genes}. All samples in JNLPBA and GENETAG have been converted into ASCII format and are annotated at the word level.

\begin{table}[!htb]
\centering
\footnotesize
\begin{center}
\begin{tabularx}{\columnwidth}{|P{4.5em}|P{2em}|Y|Y|Y|}
\hline
\bf Dataset & \bf Split & \bf \# samples & \bf \# entities & \bf \# entity types\\ \hline
\multirow{3}{*}{Bio-ID} & train & 50K$|$38K & 93K$|$90K & \multirow{3}{*}{6} \\ \cline{2-4}
& dev & 4K & 9K & \\ \cline{2-4}
& test & 14K & 30K & \\ \hline
\multirow{3}{*}{JNLPBA} & train & 31K$|$15K & 41K$|$42K & \multirow{3}{*}{5} \\ \cline{2-4}
& dev & 4K & 10K & \\ \cline{2-4}
& test & 4K & 9K & \\ \hline
\multirow{3}{*}{GENETAG} & train & 20K$|$12K & 15K$|$15K & \multirow{3}{*}{1} \\ \cline{2-4}
& dev & 3K & 4K & \\ \cline{2-4}
& test & 5K & 6K & \\ \hline
\end{tabularx}
\end{center}
\vspace{-1em}
\caption{\label{bio-dataset-stats} Dataset statistics. For training sets, the first number is the value of the dataset used for byte NN models, and the second number is the value of the dataset used for all other models.}
\end{table}

For the byte NN models, we extract overlapping samples from the original training set to collect more data for our models to train with; these additional samples are to compensate for semantic information that is usually derived from pre-trained word embeddings. For the byte NN models, we extract training samples of 150 bytes from all datasets, long enough to encompass most of the tagged entities in the training data. To extract multiple samples from an original data sample, we right-shift by 75 bytes to collect the next 150-byte sample, thereby producing new samples with some overlapping content. We experiment with extracting samples using different stride lengths; a stride of 75 bytes generally improves model performance over using samples with no overlap, while also keeping the training time reasonable. The overlapping samples in the new training set are constrained to not start or end in the middle of an entity. We also break up samples in the development and test sets into 150-byte samples, again using a stride of 75 bytes to gather the next sample; we then follow \citet{gillick2015multilingual}'s method of using overlapping samples to capture possible entities that occur at the boundary of a sample and then re-combining samples to get rid of the overlapped portions. 
%only taking predictions that end in the latter half of each 150-byte sample (except the first), in order to deal with entities that span samples.

For all other models, we pass in the original training, development, and test data without additional extraction of samples. The word NN model implementation we use takes the longest sample in the entire dataset and pads all samples to the max sample length.

\section{Methods}
We compare variations of the byte-level model with two word-level models for each dataset, and also include state-of-the-art results. For the NN models, we take sentences from 10\% of the files in the Bio-ID dataset and JNLPBA dataset and 10\% of the sentences in the GENETAG dataset to be the development sets. Our NN models learn to predict IOBES tag outputs for each byte.\footnote{IOBES and IOB are schemes for tagging parts of entities\label{iobfoot}} The IOBES and IOB schemes are similar in terms of effectiveness \cite{Reimers}; \cite{collobert2011natural} choose IOBES for expressiveness. Our NN model is relatively large, and we believe the amount of network parameters would allow us to use the more expressive scheme at a negligible cost.
% ; previous works have shown the effectiveness of using IOBES tags over IOB tags \cite{ma2016end,lample2016neural}.

\subsection{Word CRF model}
NERSuite\footnote{http://nersuite.nlplab.org/} is a CRF-based NER system that uses tokenization, lemmatization, POS-tagging, and chunking as features to tag tokens in a sequence. 
%We do not use any dictionary features or alter any default parameters. 
For each dataset, we train a NERSuite model on the training and development sets and tag each word in a sequence with an IOB tag.$^{\ref{iobfoot}}$

\subsection{Word-level NN model}
\citet{ma2016end} presents a state-of-the-art NER model that takes words as input and outputs IOBES tag predictions for each word. The BLSTM-CRF architecture uses character embeddings from convolutional neural network (CNN) layers concatenated with pre-trained word embeddings as features. For the Bio-ID dataset, we also use NERSuite's tokenizer to tokenize the data before passing it to the word-level NER model; this tokenization makes the model consistent with the tokenized JNLPBA and GENETAG datasets, even though the model thus relies on tokenization heuristics.\footnote{Without tokenization, the $F_1$ scores on the word-level NER model for Bio-ID are about 30\% lower, because many ``tokens'' do not have known word embeddings.} We use \citet{Reimers}'s word-level NER implementation.

\subsection{Byte-level NN model}
\subsubsection{Features}
All of our byte-level model variations use a subset of four features: byte embeddings, BPE embeddings, pre-trained BPE embeddings, and pre-trained word embeddings. Byte embeddings and BPE embeddings are trained in conjunction with the model. Pre-trained word embeddings\footnote{http://bio.nlplab.org/} are trained on PubMed abstracts and PubMed Central full texts, and pre-trained BPE embeddings are trained only on the latter. All pre-trained embeddings are derived from a skip-gram model \cite{mikolov2013distributed}.
%$^{\ref{sup}}$
For each byte in the input sequence, we concatenate all feature embeddings for the byte. When BPE or word features span multiple bytes, the same feature is repeated across bytes. We do a simple whitespace tokenization to decide which words (and subsequently, subwords) to get embeddings for, to keep our model free of manually-crafted tokenization rules.\footnote{We use whitespace tokenization to be compatible with the specific implementation of the BPE algorithm we use, though the general BPE algorithm could also be applied over all bytes without tokenization.}

We find that our model is slightly better when we use BPE subword tokens generated from the full PubMed Central text versus from the training data. Additionally, pre-training embeddings for BPE subword tokens improves performance. Our initial experiments also show that when using BPE features in our model, running the BPE algorithm with 5K merge operations produces the best results; when using BPE embedding features, running the BPE algorithm with 50K merge operations and then generating 100-dimensional pre-trained BPE embeddings produces the best results. In our reported results, we always use the prior configurations. Unless otherwise stated, the byte NN model with byte embeddings and pre-trained BPE embeddings as features is the general ``byte NN'' model that we report results for. These features, along with the general byte CNN-BLSTM-CRF architecture, produce the best results.

\begin{figure}[thb]
\includegraphics[scale=0.2]{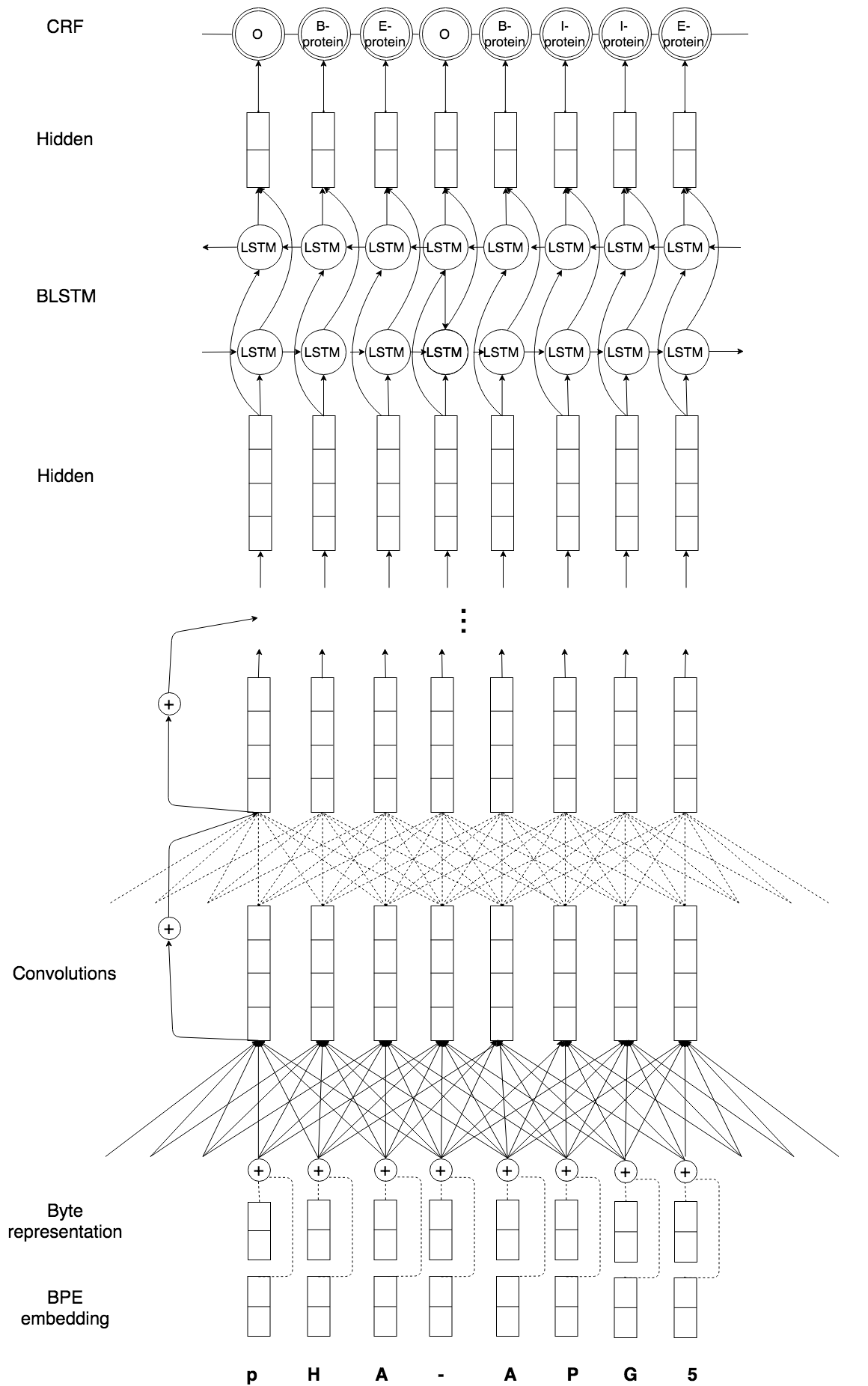}
\caption{Byte NN architecture. Dashed lines indicate dropout.}
\label{arch}
\vspace{-1em}
\end{figure}

\subsubsection{Architecture}
% Our early experiments show that a stack of 20 CNN layers with residual connections between each layer is able to extract helpful information for our byte NER model. 
The model starts with a stack of 20 CNN layers with residual connections between each layer. Following the pattern of effective neural NER architectures, the CNN stack is followed by a BLSTM layer and then a CRF layer, with hidden layers in between, as shown in Figure \ref{arch}. Our preliminary experiments indicate that a stack of CNNs and residual connections are necessary for our byte-level models to reach comparable performance with the word-level models.
% $^{\ref{sup}}$

% Other works typically concatenate the CNN-derived character representation to pre-trained word embeddings, and then pass the concatenation further along the network. 
We find that passing the pre-trained embeddings through the entire CNN-BLSTM-CRF network and also allowing the embeddings to be fine-tuned through the CNN layers improve the overall scores. Additional dropout \cite{hinton2012improving} of embeddings and after each CNN layer further improves model performance. We also incorporate byte-dropout \cite{gillick2015multilingual}, a technique that makes the model more robust to noise by randomly replacing a percentage of input bytes with a special DROP symbol.

\subsubsection{Hyperparameters}
\label{sec:supplemental}
For the byte NN model, we use dropout with a rate of 0.5, byte-dropout with a rate of 0.3, a learning rate of 0.0001 with Adam, and a mini batch size of 256 samples. The pre-trained word embeddings are 200-dimensional embeddings, and the pre-trained BPE embeddings are 100-dimensional embeddings. We use CNNs with 250 filters, a filter size of 7 bytes, a filter stride of 1 byte, and a ReLu activation function.
The BLSTM layer also has 250 units and uses a tanh activation function. We run the byte NER models for 300 epochs. Non-pre-trained embeddings are initialized with a random uniform distribution [-0.05, 0.05].

BPE embeddings are 100-dimensional embeddings and are trained for 10 iterations using the skip gram model with a window size of 5 tokens.

The word NN model has a mini batch size of 32 samples, a clipnorm of 1, an output dropout of 0.5, a recurrent dropout of 0.5, a default learning rate of 0.002 with Nadam. It uses a CNN layer with 25 filters, a filter size of 7 characters, a filter stride of 1 character and a ReLu activation function to get character embeddings. Additional features (tokens and casing) have the default dimensions of 10. The BLSTM layer has  200 units and uses a tanh activation function. The model is run for 100 epochs without early stopping.

\section{Results}
\begin{table}[!htb]
\centering
\footnotesize
\begin{center}
\begin{tabularx}{\columnwidth}{|P{4.5em}||Y|Y|Y||P{5.1em}|}
\hline
\textbf{Entity type} & \textbf{Word CRF} & \textbf{Word NN} & \textbf{Byte NN} & \textbf{Best @ BioCreative VI} \\ \hline
cell type or line & \textbf{72.23} & 71.78 & 71.81 & 74.4 \\ \hline
cellular component & 56.55 & \textbf{63.98} & 58.62 & 57.9 \\ \hline
organisms or species & 78.43 & 79.16 & \textbf{81.97} & 83.4 \\ \hline
protein or gene & 70.79 & 76.00 & \textbf{79.31} & 73.4 \\ \hline
small molecule & 67.34 & \textbf{76.01} & 65.45 & 66.8 \\ \hline
tissues or organs & 62.79 & \textbf{66.31} & 62.91 & 64.3 \\ \hline
Total & 69.72 & 74.25 & \textbf{74.73} & - \\ \hline
\end{tabularx}
\end{center}
\vspace{-1em}
\caption{\label{bio-id} $F_1$ scores across Bio-ID categories. Best entity results, excluding last column, are bolded.}

\centering
\footnotesize
\begin{center}
\begin{tabularx}{\columnwidth}{|P{3.4em}||Y|Y|Y||P{3.7em}|}
\hline
\textbf{Entity type} & \textbf{Word CRF} & \textbf{Word NN} & \textbf{Byte NN} & \textbf{(Gridach, 2017)} \\ \hline
cell type & 71.45 & \textbf{74.66} & 70.90 & - \\ \hline
cell line & 54.90 & \textbf{60.17} & 56.76 & - \\ \hline
dna & 67.29 & \textbf{70.36} & 67.25 & - \\ \hline
protein & 70.12 & \textbf{75.31} & 70.42 & - \\ \hline
rna & 67.51 & \textbf{68.27} & 68.02 & - \\ \hline
Total & 69.09 & \textbf{73.53} & 69.26 & 75.87 \\ \hline
\end{tabularx}
\end{center}
\vspace{-1em}
\caption{\label{jnlpba} $F_1$ scores across JNLPBA categories. Best entity results, excluding last column, are bolded.}

\centering
\footnotesize
\begin{center}
\begin{tabularx}{\columnwidth}{|P{5em}||Y|Y|Y||P{3.4em}|}
\hline
\textbf{Entity type} & \textbf{Word CRF} & \textbf{Word NN} & \textbf{Byte NN} & \textbf{(Gridach, 2017)} \\ \hline
protein/gene  & 84.61 & \textbf{89.45} & 85.54 & 89.46 \\ \hline
\end{tabularx}
\end{center}
\vspace{-1em}
\caption{\label{genetag} $F_1$ scores across GENETAG protein/genes. Best entity results, excluding last column, are bolded.}

\end{table}

Table \ref{bio-id} compares the $F_1$ scores of entities in the Bio-ID dataset tagged by our models. The byte NN model is better at finding \textit{cell type or lines}, \textit{organisms or species}, and \textit{protein or genes} than the word NN model. 
% The byte NN model also outperforms the best model submitted to BioCreative VI Track 1 for some categories, and comes close in others. 
We examine the fact that the word NN model has an $F_1$ score 10\% higher than that of other models for \textit{small molecules}. Although a large number (55\%) of the entities in the Bio-ID dataset are \textit{protein and genes}, we find that the proportion of \textit{small molecules} mistaken for \textit{protein or genes} is higher than that of other entities mistaken for \textit{protein or genes}. Looking at overall sequences of words may be necessary for more accurate identification of \textit{small molecules}. 

The best model submitted to BioCreative VI Track 1 uses a word-level CRF-based approach, along with preprocessing and heuristics \cite{kaewphan2017turkunlp}. The byte NN model outperforms all other models for \textit{protein or gene} categories; importantly, the byte NN model is the \textbf{only fully learned model} that does not rely on heuristics for tokenization and other processing.

Tables \ref{jnlpba} and \ref{genetag} show that the byte-level model does not beat the word-level model on the JNLPBA and GENETAG datasets. Because the annotation of JNLPBA and GENETAG were explicitly constrained to words, we believe they do not serve as useful bases for our exploration of byte-level models. Our initial results on these datasets indicate that fully end-to-end byte-level models may be more suitable for entities whose spans do not align with word spans.

\begin{table}[!htb]
\centering
\footnotesize
\begin{center}
\begin{tabularx}{\columnwidth}{|P{4.5em}||Y|Y|P{2.5em}|P{2.5em}|P{2.5em}|}
\hline
\textbf{Entity type} & \textbf{Bytes} & \textbf{BPE} & \textbf{Pre-trained BPE} & \textbf{Pre-trained word} & \textbf{Bytes + Pre-trained BPE} \\ \hline
cell type or line & 67.59 & 69.15 & 70.77 & 62.01 & \textbf{71.81} \\ \hline
cellular component & 54.25 & 57.30 & 58.52 & 50.31 & \textbf{58.62} \\ \hline
organisms or species & 79.56 & 80.61 & \textbf{83.05} & 74.04 & 81.97 \\ \hline
protein or gene & 73.60 & 76.51 & 77.91 & 50.52 & \textbf{79.31} \\ \hline
small molecule & 57.77 & 61.83 & \textbf{65.46} & 55.39 & \textbf{65.45} \\ \hline
tissues or organs & 60.46 & 63.35 & \textbf{64.44} & 54.97 & 62.91 \\ \hline
Total & 69.41 & 72.37 & 73.97 & 54.96 & \textbf{74.73} \\ \hline
\end{tabularx}
\end{center}
\vspace{-1em}
\caption{\label{feature-comparison} $F_1$ scores across Bio-ID categories for byte NN model. Columns are feature(s) used. Best entity results are bolded.}
\end{table}

We also look at the effect of byte, BPE, and word features in Table \ref{feature-comparison}. Previous works have shown that pre-trained word embeddings are important features for word-level NER models; we find that they are less useful for byte-level models. For a consistent feature set across bytes, contiguous bytes belonging to the same word have the same word feature. This repetition of information may diminish the effectiveness of word embeddings in the byte-level models. However, even though we repeat BPE features in the same way, table \ref{feature-comparison} shows that BPE features are useful. Because the Bio-ID dataset is dominated by \textit{protein or genes}, the byte NN model trained on byte and pre-trained BPE embeddings has a higher overall micro-$F_1$ score than the byte NN model that only uses pre-trained BPE embeddings. With these results, we emphasize that BPE features are useful subword information for NER at the byte-level.

\section{Conclusion}
Our initial experiments on the byte-level NER models across datasets motivate these models as a useful end-to-end alternative for entities that naturally exist at the subword level.
% At the current state, byte-level models provide superior performance in some instances but do not consistently outperform state-of-the-art word-level models on standard NER datasets; however, we show that byte-level models are automatic models that operate without manual heuristics and produce comparable or better results for entities that are not constrained to word-level annotations. 
Further investigations into byte-level models could help facilitate more precise byte-level annotation schemes for the biomedical domain.

\section*{Acknowledgments}
We would like to thank Jos\'e-Luis Ambite, Scott Miller, Aram Galstyan, Ryan Gabbard, as well as all the anonymous reviewers for their invaluable advice regarding this work. 

\bibliography{emnlp2018}

\begin{thebibliography}{}
\expandafter\ifx\csname natexlab\endcsname\relax\def\natexlab#1{#1}\fi

\bibitem[{Abujabal and Gaspers(2018)}]{abujabal2018neural}
Abdalghani Abujabal and Judith Gaspers. 2018.
\newblock Neural named entity recognition from subword units.
\newblock {\em arXiv preprint arXiv:1808.07364\/} .

\bibitem[{Arighi et~al.(2017)Arighi, Hirschman, Lemberger, Bayer, Liechti,
  Comeau, and Wu}]{arighi2017bioid}
Cecilia Arighi, Lynette Hirschman, Thomas Lemberger, Samuel Bayer, Robin
  Liechti, Donald Comeau, and Cathy Wu. 2017.
\newblock Bio-id track overview.
\newblock In {\em Proceedings of BioCreative VI Workshop\/}. BioCreative, pages
  14--19.

\bibitem[{Chiu and Nichols(2015)}]{chiu2015named}
Jason~PC Chiu and Eric Nichols. 2015.
\newblock Named entity recognition with bidirectional lstm-cnns.
\newblock {\em arXiv preprint arXiv:1511.08308\/} .

\bibitem[{Collobert et~al.(2011)Collobert, Weston, Bottou, Karlen, Kavukcuoglu,
  and Kuksa}]{collobert2011natural}
Ronan Collobert, Jason Weston, L{\'e}on Bottou, Michael Karlen, Koray
  Kavukcuoglu, and Pavel Kuksa. 2011.
\newblock Natural language processing (almost) from scratch.
\newblock {\em Journal of Machine Learning Research\/} 12(Aug):2493--2537.

\bibitem[{Gage(1994)}]{gage1994anew}
Philip Gage. 1994.
\newblock A new algorithm for data compression.
\newblock {\em C Users J\/} 12(2):23--38.

\bibitem[{Gillick et~al.(2015)Gillick, Brunk, Vinyals, and
  Subramanya}]{gillick2015multilingual}
Dan Gillick, Cliff Brunk, Oriol Vinyals, and Amarnag Subramanya. 2015.
\newblock Multilingual language processing from bytes.
\newblock {\em arXiv preprint arXiv:1512.00103\/} .

\bibitem[{Gridach(2017)}]{gridach2017character}
Mourad Gridach. 2017.
\newblock Character-level neural network for biomedical named entity
  recognition.
\newblock {\em Journal of biomedical informatics\/} 70:85--91.

\bibitem[{Hinton et~al.(2012)Hinton, Srivastava, Krizhevsky, Sutskever, and
  Salakhutdinov}]{hinton2012improving}
Geoffrey~E Hinton, Nitish Srivastava, Alex Krizhevsky, Ilya Sutskever, and
  Ruslan~R Salakhutdinov. 2012.
\newblock Improving neural networks by preventing co-adaptation of feature
  detectors.
\newblock {\em arXiv preprint arXiv:1207.0580\/} .

\bibitem[{Kaewphan et~al.(2017)Kaewphan, Mehryary, Hakala, Salakoski, and
  Ginter}]{kaewphan2017turkunlp}
Suwisa Kaewphan, Farrokh Mehryary, Kai Hakala, Tapio Salakoski, and Filip
  Ginter. 2017.
\newblock Turkunlp entry for interactive bio-id assignment.
\newblock In {\em Proceedings of the BioCreative VI Workshop\/}. BioCreative VI
  Workshop Proceedings.

\bibitem[{Kim et~al.(2004)Kim, Ohta, Tsuruoka, Tateisi, and
  Collier}]{kim2004introduction}
Jin-Dong Kim, Tomoko Ohta, Yoshimasa Tsuruoka, Yuka Tateisi, and Nigel Collier.
  2004.
\newblock Introduction to the bio-entity recognition task at jnlpba.
\newblock In {\em Proceedings of the international joint workshop on natural
  language processing in biomedicine and its applications\/}. Association for
  Computational Linguistics, pages 70--75.

\bibitem[{Kuru et~al.(2016)Kuru, Can, and Yuret}]{kuru2016charner}
Onur Kuru, Ozan~Arkan Can, and Deniz Yuret. 2016.
\newblock Charner: Character-level named entity recognition.
\newblock In {\em Proceedings of COLING 2016, the 26th International Conference
  on Computational Linguistics: Technical Papers\/}. pages 911--921.

\bibitem[{Lample et~al.(2016)Lample, Ballesteros, Subramanian, Kawakami, and
  Dyer}]{lample2016neural}
Guillaume Lample, Miguel Ballesteros, Sandeep Subramanian, Kazuya Kawakami, and
  Chris Dyer. 2016.
\newblock Neural architectures for named entity recognition.
\newblock {\em arXiv preprint arXiv:1603.01360\/} .

\bibitem[{Lyu et~al.(2017)Lyu, Chen, Ren, and Ji}]{lyu2017long}
Chen Lyu, Bo~Chen, Yafeng Ren, and Donghong Ji. 2017.
\newblock Long short-term memory rnn for biomedical named entity recognition.
\newblock {\em BMC bioinformatics\/} 18(1):462.

\bibitem[{Ma and Hovy(2016)}]{ma2016end}
Xuezhe Ma and Eduard Hovy. 2016.
\newblock End-to-end sequence labeling via bi-directional lstm-cnns-crf.
\newblock {\em arXiv preprint arXiv:1603.01354\/} .

\bibitem[{Mikolov et~al.(2013)Mikolov, Sutskever, Chen, Corrado, and
  Dean}]{mikolov2013distributed}
Tomas Mikolov, Ilya Sutskever, Kai Chen, Greg~S Corrado, and Jeff Dean. 2013.
\newblock Distributed representations of words and phrases and their
  compositionality.
\newblock In {\em Advances in neural information processing systems\/}. pages
  3111--3119.

\bibitem[{Misawa et~al.(2017)Misawa, Taniguchi, Miura, and
  Ohkuma}]{misawa2017character}
Shotaro Misawa, Motoki Taniguchi, Yasuhide Miura, and Tomoko Ohkuma. 2017.
\newblock Character-based bidirectional lstm-crf with words and characters for
  japanese named entity recognition.
\newblock In {\em Proceedings of the First Workshop on Subword and Character
  Level Models in NLP\/}. pages 97--102.

\bibitem[{Reimers and Gurevych(2017)}]{Reimers}
Nils Reimers and Iryna Gurevych. 2017.
\newblock \href{http://aclweb.org/anthology/D17-1035}{Reporting score
  distributions makes a difference: Performance study of lstm-networks for
  sequence tagging}.
\newblock In {\em Proceedings of the 2017 Conference on Empirical Methods in
  Natural Language Processing (EMNLP)\/}. Copenhagen, Denmark, pages 338--348.
\newblock
  \href{http://aclweb.org/anthology/D17-1035}{http://aclweb.org/anthology/D17-1035}.

\bibitem[{Sennrich et~al.(2015)Sennrich, Haddow, and
  Birch}]{sennrich2015neural}
Rico Sennrich, Barry Haddow, and Alexandra Birch. 2015.
\newblock Neural machine translation of rare words with subword units.
\newblock {\em arXiv preprint arXiv:1508.07909\/} .

\bibitem[{Sutskever et~al.(2014)Sutskever, Vinyals, and
  Le}]{sutskever2014sequence}
Ilya Sutskever, Oriol Vinyals, and Quoc~V Le. 2014.
\newblock Sequence to sequence learning with neural networks.
\newblock In {\em Advances in neural information processing systems\/}. pages
  3104--3112.

\bibitem[{Tanabe et~al.(2005)Tanabe, Xie, Thom, Matten, and
  Wilbur}]{tanabe2005genetag}
Lorraine Tanabe, Natalie Xie, Lynne~H Thom, Wayne Matten, and W~John Wilbur.
  2005.
\newblock Genetag: a tagged corpus for gene/protein named entity recognition.
\newblock {\em BMC bioinformatics\/} 6(1):S3.

\bibitem[{Verga et~al.(2018)Verga, Strubell, and
  McCallum}]{verga2018simultaneously}
Patrick Verga, Emma Strubell, and Andrew McCallum. 2018.
\newblock Simultaneously self-attending to all mentions for full-abstract
  biological relation extraction.
\newblock {\em arXiv preprint arXiv:1802.10569\/} .

\end{thebibliography}
\bibliographystyle{acl_natbib}

% \appendix

% \section{Supplemental Material}
% \label{sec:supplemental}
% For the byte NN model, we use dropout with a rate of 0.5, byte-dropout with a rate of 0.3, a learning rate of 0.0001 with Adam, and a mini batch size of 256 samples. The pre-trained word embeddings are 200-dimensional embeddings, and the pre-trained BPE embeddings are 100-dimensional embeddings. We use CNNs with 250 filters, a filter size of 7 bytes, a filter stride of 1 byte, and a ReLu activation function.
% The BLSTM layer also has 250 units and uses a tanh activation function. We run the byte NER models for 300 epochs. Non-pre-trained embeddings are initialized with a random uniform distribution [-0,05, 0.05].

% BPE embeddings are 100-dimensional embeddings and are trained for 10 iterations using the skip gram model with a window size of 5 tokens.

% The word NN model has a mini batch size of 32 samples, a clipnorm of 1, an output dropout of 0.5, a recurrent dropout of 0.5, a default learning rate of 0.002 with Nadam. It uses a CNN layer with 25 filters, a filter size of 7 characters, a filter stride of 1 character and a ReLu activation function to get character embeddings. Additional features (tokens and casing) have the default dimensions of 10. The BLSTM layer has  200 units and uses a tanh activation function. The model is run for 100 epochs without early stopping.

\end{document}